%% file: acl_latex.tex
\pdfoutput=1

\documentclass[11pt]{article}

\usepackage[]{acl}

\usepackage{times}
\usepackage{latexsym}

\usepackage[T1]{fontenc}

\usepackage[utf8]{inputenc}

\usepackage{microtype}

\usepackage{inconsolata}

\usepackage{graphicx}

\usepackage{tabularx}
\usepackage{lipsum}
\usepackage{makecell}
\usepackage{hhline}
\usepackage{booktabs}
\usepackage{hyperref}
\usepackage{natbib}
\usepackage{graphicx}
\usepackage{subcaption}
\usepackage{multirow}
\usepackage{longtable}
\usepackage{enumitem}
\usepackage{caption}
\usepackage{xcolor}
\usepackage{enumitem}
\usepackage{amssymb}
\usepackage{titlesec}
\titlespacing*{\paragraph}{0pt}{0pt}{1ex plus 0.2ex minus 0.2ex}

\title{Instructions for *ACL Proceedings}

\author{First Author \\
  Affiliation / Address line 1 \\
  Affiliation / Address line 2 \\
  Affiliation / Address line 3 \\
  \texttt{email@domain} \\\And
  Second Author \\
  Affiliation / Address line 1 \\
  Affiliation / Address line 2 \\
  Affiliation / Address line 3 \\
  \texttt{email@domain} \\}

\title{HateCOT: An Explanation-Enhanced Dataset for Generalizable Offensive Speech Detection via Large Language Models}

\author{
  Huy Nghiem  \\
  University of Maryland \\
  \texttt{nghiemh@umd.edu} \\ 
  \\\And
  Hal Daumé III\\
  University of Maryland \& Microsoft Research \\
  \texttt{hal3@umd.edu} \\}
  
\defcitealias{mathew2021hatexplain}{HateXplain}
\defcitealias{elsherief2021latent}{Latent\_Hate}
\defcitealias{rottger-etal-2021-hatecheck}{HateCheck}

\newcommand{\defaultcolor}{black}
\newcommand{\secondcolor}{red}

\newcommand{\clt}[1]{\textcolor{\defaultcolor}{#1}}
\newcommand{\hlt}[1]{\textcolor{\secondcolor}{#1}}

\begin{document}
\maketitle
\begin{abstract}
    \hlt{\textbf{Warning:} \textit{This paper contains examples of very offensive material.}}
    The widespread use of social media necessitates reliable and efficient detection of offensive content to mitigate harmful effects. Although sophisticated models perform well on individual datasets, they often fail to generalize due to varying definitions and labeling of "offensive content." In this paper, we introduce \textit{HateCOT}, an English dataset with over 52,000 samples from diverse sources, featuring explanations generated by GPT-3.5-Turbo and curated by humans. We demonstrate that pretraining on \textit{HateCOT} significantly enhances the performance of open-source Large Language Models on three benchmark datasets for offensive content detection in both zero-shot and few-shot settings, despite differences in domain and task. Additionally, \textit{HateCOT} facilitates effective K-shot fine-tuning of LLMs with limited data and improves the quality of their explanations, as confirmed by our human evaluation. Our repository is available at \url{https://github.com/hnghiem-usc/hatecot}
.
\end{abstract}

\input{intro}

\input{related_work}

\input{data}
\input{experiment}

\input{qa}

\input{limitation}

\bibliography{anthology, custom}

\input{appendix}

\end{document}

%% file: intro.tex
\section{Introduction}
As social media has become indispensable to modern discourse, this channel of communication has amplified the propagation of offensive content. Speech that promotes hateful sentiments thrives on such platforms, leading to real and significant harm on their audience \cite{giachanou2020battle, saha2019prevalence}. However, ``offensive content'' is still a contested construct, as what is and is not allowed varies by platform. In research, different approaches analyze semantically similar but still distinct concepts: \textit{Cyber-bullying, Toxicity, Sexist, Racist, Hate etc.} \cite{poletto2021resources, fortuna2021well, nghiem-2024, nguyen2023towards}, further highlighting this contestedness.

Compounding the challenge, reliable detection of offensive content typically requires significant amounts of data. Sophisticated models tend to be data-hungry, and the process of curating a dataset tailored to a specific use case can be costly, time-consuming, and emotionally challenging for annotators \cite{founta2018large, toraman2022large}. The typical pipeline consists of collecting samples based on topic-relevant key words, then recruiting either crowdworkers or experts to annotate data before developing classification models \cite{paullada2021data}. Each step incurs investment and may inject subtle idiosyncrasies \clt{proportionate to the size of the} downstream dataset, further limiting transferable usefulness to related tasks \cite{fortuna2021well}. \clt{The size of a dataset also does not necessarily guarantee cross-domain transferrability \cite{poletto2021resources, fortuna2021well}.}

In practical settings, users may desire transparency from social media platforms. Therefore, the ability to provide human-understandable justification based on platform-specific policy becomes an attractive feature for content moderation. Nevertheless, current techniques often still fail to offer intuitive explanatory signals \cite{yadav2023hate, babaeianjelodar2022interpretable, ibrahim2022explainable}.

In this work, we attempt to simultaneously reduce the cost of \textit{data curation}, enhance \textit{cross-dataset generalization}, and address the necessity of \textit{explainable decisions} for offensive content detection. Specifically, our main contributions are: 
\begin{enumerate}[itemsep=0pt, labelsep=5pt]
    \item We release \textbf{HateCOT} (\textit{Hate-related Chains-of-Thought}), a dataset of over $52,000$ samples consisting of input text, a hate speech label, and an explanation for that label. This corpus is constructed by merging eight datasets with explanations created by using GPT-3.5-Turbo to augment human annotations.

    \item We demonstrate the benefits of using \textit{HateCOT} as a pretraining corpus before finetuning on a target domain. Empirical results across 3 datasets show that open-source Large Language Models (LLMs) can effectively leverage definitions to adapt to new tasks using zero-shot and few-shot settings via finetuning and in-context learning .  
    
    \item We assess the quality of explanations generated by our finetuned models with respect to the criteria described in their corresponding definitions. These insights showcase LLM-generated explanations as a means to enhance transparency in content moderation.
\end{enumerate}

%% file: related_work.tex
\section{Related Works}
\subsection{Offensive Speech Detection}
Offensive speech detection has attracted considerable interest from the research community. Earlier approach typically investigated coarse-grained labels (e.g. \textit{Hate} vs \textit{Not Hate}) while subsequent efforts explored more diverse facets of offensive speech at higher granularity \cite{founta-specia-2021-survey, poletto2021resources, vidgen2020directions}. Increasingly more advanced models emerged over time with the diversity of datasets.

Cross-domain generalization, however, still remains a relevant challenge in the area. \citet{fortuna2021well} found empirically that cross-dataset transference is highly dependent on semantic similarity between their label spaces. Recent works have pretrained Transformer-based models, such as HateBERT and fBERT, on specialized corpora to enhance generalization to various levels of success \cite{caselli2021hatebert, sarkar2021fbert}.

\subsection{LLMs in Offensive Speech Classification} 
\citet{zampieri2023offenseval} assessed a range of open-sourced LLMs on zero-shot prompting on the OffensEval task and found their performance trailing by a wide margin behind trained existing finetuned BERT-based systems. \citet{chiu2021detecting} and \citet{han2022designing} used the proprietary GPT-3 on a set of different datasets and noted that informative contexts and examples could boost the model's performance to competitive levels on a different set of data. Similarly, \citet{roy2023probing} found that adding explanation to pipeline could result in 10 to 20\% boost in performance of LLMs over baselines. \citep{yang2023hare}'s study found that training LLMs with step-by-step reasoning grounded by annotations could improve  predictive power. 

Pretrained language models have exhibited remarkable ability in text generation \cite{celikyilmaz2020evaluation}. Recent large-size LLMs such as GPT-3 and later models are capable of astoundingly fluent, convincing and knowledge-infused outputs \cite{zhang2023survey}. LLMs with hundreds of billions of parameters even exhibit reasoning capabilities \cite{wei2022chain, wei2021finetuned}, leading to a flurry of research on prompting techniques to harness their prowess, such as Chain-of-thought (COT), Tree-of-thought etc. \citep{yao2023tree, diao2023active}. An interesting line of research leverages LLMs to efficiently generate high volumes of synthetic data for tasks with training resource is scarce \cite{puri2020training, bao2023synthetic, whitehouse2023llm}. We build upon these works to construct a dataset that can induce smaller LLMs to efficiently adapt to new categories of offensive content by leveraging their provided definitions.

%% file: data.tex
\section{Building \textit{HateCOT}}
We first describe the process to identify the candidate datasets from literature, and the procedure to obtain annotation-guided explanations from these samples (Section \ref{sec:data}). We then perform a set of validation experiments to optimize the data's parameters for downstream  tasks before augmenting our corpus to its eventual size (Section \ref{sec:optim}).

\subsection{Data Selection}\label{sec:data}


\paragraph{Datasets for Training.}  We use the following criteria to filter existing corpora related to offensive speech detection: 
\begin{itemize}[topsep=0pt, partopsep=0pt,itemsep=0pt, parsep=0pt,label=$\triangleright$]
    \item \textit{Size}: datasets should contain more than 5,000 samples to ensure adequate size for subsequent sampling.
    \item \textit{Label}: datasets should contain diverse label space that address different facets of offensive language. Both neutral and non-neutral categories should be included for parity. 
    \item \textit{Definition}: each dataset should have the associated definitions with each label available (Figure \ref{apx:train_def}). This criteria is important to generate informative explanations. 
    \item \textit{Target / Rationale}: the dataset should provide the \textit{a)} targets and/or \textit{b)} rationales,  \clt{which are fragments of free texts by human annotators to convey some understanding of the corresponding post. Figure \ref{fig: rationale} shows several examples of rationales, demonstrating their unsuitability to be used as explanations in their native format.}
\end{itemize}
These criteria significantly reduce the number of eligible candidates since many do not provide the required annotation on target and definition. Table \ref{tab:train_data} lists the 8 selected datasets.
\paragraph{Datasets for Evaluation.} 
Using similar criteria, we select 3 additional datasets with different label spaces and definitions for downstream testing (details in Table \ref{tab:test_data}).

\textbf{\textit{HateCheck}} was created by \citet{rottger-etal-2021-hatecheck} with the explicit goal of evaluating hate speech detection models. 10 trained annotators labelled the dataset using a binary schema: \textit{Hateful} and \textit{Non-hateful}, with the reported inter-annotator agreement coefficient to be 0.93.

\textbf{\textit{HateXplain}} was primarily collected from Twitter and the Gab platform \cite{mathew2021hatexplain}. In addition to the labels \textit{Hate, Offensive, Normal}, annotators also provide justification for their selection by highlighting the span of tokens, called rationales, that contribute to their decision.

\textbf{\textit{Latent\_Hate}} was created on the premise that offensive speech classifiers tend to bias towards covert negative sentiment \cite{elsherief2021latent}. After discarding the augmented portion from the Social\_Bias dataset to avoid contamination, 22,584 samples collected from Twitter remained. This dataset contains 3 coarse-grained labels \textit{Not Hate, Explicit Hate, Implicit Hate}, while a subset contains 6 fine-grained categories, which we refer to as \textbf{Implicit\_Hate} in subsequent test regimens.

\begin{table*}[ht]
\begin{subtable}[t]{\textwidth} 
\footnotesize
\renewcommand{\arraystretch}{1.35}  
\newcolumntype{C}[1]{>{\centering\arraybackslash}p{#1}}
\newcolumntype{R}[1]{>{\raggedleft\arraybackslash}p{#1}} 

{\begin{tabular}{p{2.75cm}R{1.25cm}R{1cm}p{1cm}C{0.75cm}C{0.75cm}p{5.5cm}}
\toprule  
\textbf{Dataset} & \textbf{Total Size} & \textbf{Sample Size} & \textbf{Platform} & \textbf{Target} & \textbf{Ration.} & \textbf{Selected Labels} \\
\midrule  
\citet{salminen2018anatomy}   & 137,098 & 5, 418 & Y, F    &  & \checkmark & Hateful, Neutral         \\
\citet{qian2019benchmark}    & 34,000   & 5, 034 & G                 &  & \checkmark & Not Hate, Hate           \\
\citet{sap2019social}       & 44,671   & 6, 033 & G, R, T & \checkmark & \checkmark & Not Offensive, Offensive \\
\citet{vidgen2021introducing}    & 27,494  & 6, 717 & R                  & \checkmark & \checkmark & Neutral, Person Directed Abuse, Affiliation Directed Abuse, Identity Directed Abuse \\
\citet{vidgen-etal-2021-learning} & 10,152  & 7, 209 & S               & \checkmark &  & None, Derogation, Dehumanization, Animosity, Support, Threatening                  \\
\citet{sachdeva2022measuring}    & 135,556 & 7, 272 & Y, T, R & \checkmark &  & Not Hate speech, Hate Speech                                                        \\
\citet{hartvigsen2022toxigen} & 274,186 & 7, 239 & S            & \checkmark &  & Benign, Toxic            \\
\citet{toraman2022large}     & 100,000 & 7, 215 & T              &  & \checkmark & Normal, Offensive, Hate  \\
\hline 
\textbf{Total} & - & \textbf{52, 137} & & & & \\ 
\bottomrule  
\end{tabular}}
\caption{Datasets used to create training corpus. \textit{Sample Size} denotes the number of chosen samples from corresponding dataset included in the training corpus.}

\label{tab:train_data}
\end{subtable} 

\begin{subtable}[t]{\textwidth}
\footnotesize
\renewcommand{\arraystretch}{1.35}  
\newcolumntype{C}[1]{>{\centering\arraybackslash}p{#1}}
\newcolumntype{R}[1]{>{\raggedleft\arraybackslash}p{#1}} 
{\begin{tabular}{p{2cm}R{1cm}R{0.75cm}R{0.25cm}R{0.15cm}p{1cm}C{0.7cm}C{0.7cm}p{5cm}}
\toprule  
\textbf{Dataset} & \textbf{Total Size} & \textbf{Split Ratio} & \textbf{K val} & \textbf{K test} & \textbf{Platform} & \textbf{Target} & \textbf{Ration.} & \textbf{Selected Labels} \\
\midrule  
\citetalias{rottger-etal-2021-hatecheck}              & 3,728  & 50:50 & 300 & 500 & S    & \checkmark &  & Non-hateful, Hateful                   \\
\citetalias{mathew2021hatexplain} & 20,148 & 60:40 & 200 & 400 & G, T & \checkmark & \checkmark & Normal, Offensive, Hate                \\
\citetalias{elsherief2021latent} & 19,112 & 60:40 & 200 & 400 & T    & \checkmark & \checkmark & Not Hate, Explicit Hate, Implicit Hate \\
Implicit\_Hate & 4,153 & 60:40 & - & 150 & T &  \checkmark & \checkmark &
  White Grievance, Incitement to Violence, Inferiority Language, Irony, Stereotypes and Misinformation, Threatening and Intimidation \\
  \hline
\end{tabular}}
\caption{Datasets for testing}
\label{tab:test_data}
\end{subtable} 
\caption{\textit{Sample Size} denotes the number of entries in the final corpus. \textit{Target} and \textit{Ration.} indicates the availability of annotation on Target or Rationale in the dataset. For \textit{Platform}, F: Facebook, Y: Youtube, G: Gab, R: Reddit, S: Synthetic, T: Twitter. \textit{K val} and \textit{K test} represent the number of sampler per class drawn during development of the training corpus and final testing, respectively. Full definitions in Table \ref{apx:train_def} and \ref{apx:test_def}.}
\end{table*}

\begin{figure}[t]
  \centering
    \includegraphics[width=\linewidth]{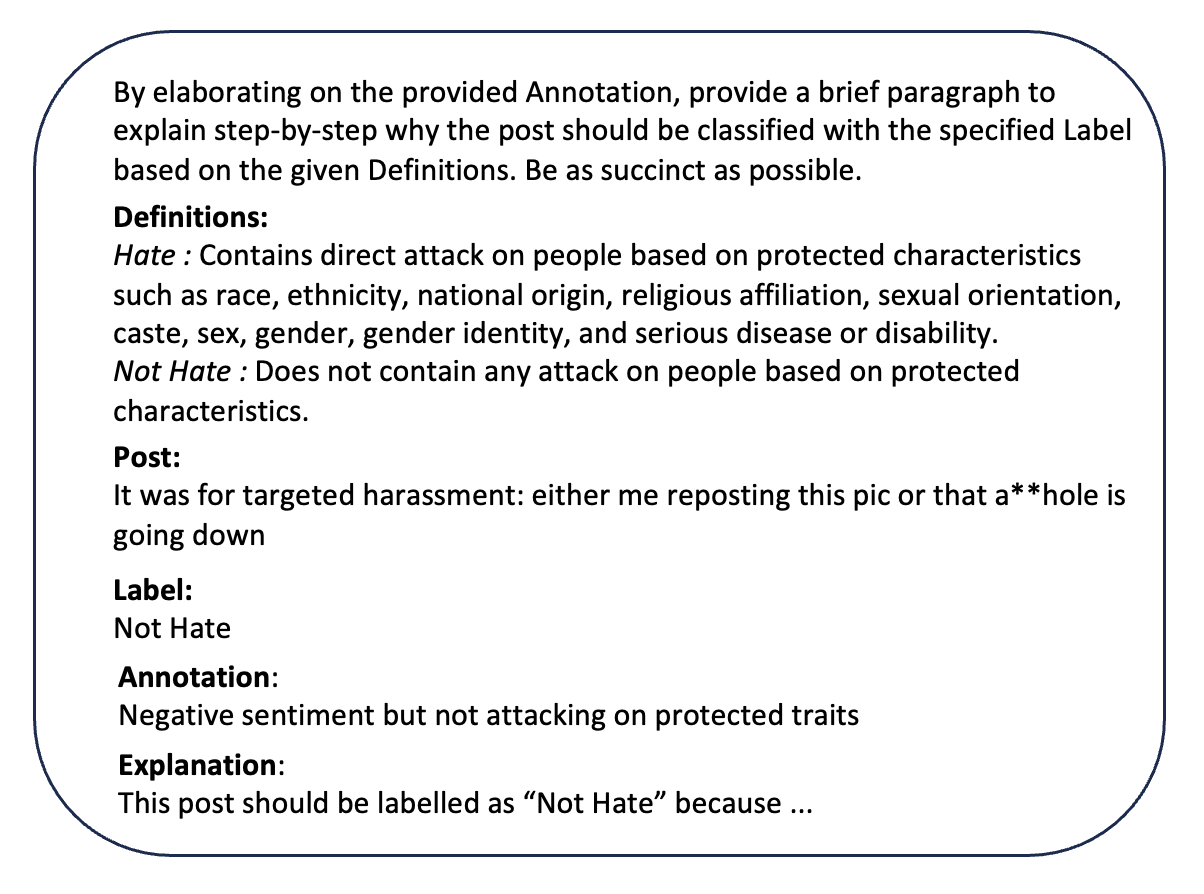}
  \caption{Template used to obtain explanations from GPT-3.5-Turbo guided by human-annotated rationales. }
  \label{fig:cot_template}
\end{figure}

 \paragraph{Obtaining Annotation-Guided Explanation.} \label{sec:exp} 
Inspired by \citet{yang2023hare}'s work that shows that GPT-3.5 could augment human-written rationales to create coherent texts that are still faithful to the original content, we use the prompt template in Figure \ref{fig:cot_template} to generate the explanation, which is guided by the available annotations on label, target, and rationale from the chosen corpora. For datasets that contain multiple annotations per sample, we select the ultimate label via majority voting and concatenate  annotations on the targets and/or rationales into a single string delimited by \textit{"|"}. 



We use GPT-3.5-Turbo, accessed via the OpenAI's API, to generate explanations due to this model's affordability and its capability to follow instructions and generate coherent outputs \citep{ye2023comprehensive, koubaa2023gpt}. For each of the 8 training datasets, we first randomly select and qualitatively analyze 20 samples to ensure the generated explanations are \textit{a) stylistically coherent}, \textit{b) consistent with the provided labels}, and \textit{c) congruent with the criteria denoted by the definitions.}  If deemed unsatisfactory, we iteratively adjust the input prompt until the quality threshold is achieved. Appendix \ref{apx:manual_review} describes this quality assurance process and the final prompts for different scenarios.

\subsection{Optimization of Synthesized Corpus}\label{sec:optim}


Extending previous works \cite{magister2022teaching, ho2022large}, we are interested in optimizing 2 parameters central to the construction of our corpus: the distribution of neutral vs. non-neutral classes in the data and the number of explanations \textit{per sample}. The former has been noted to influence predictive powers \cite{rathpisey2019handling, casula2020hate}. The latter, also referred to as \textit{degree of reasoning diversity}, could improve knowledge distillation \cite{ho2022large}. We use the open-source model \textbf{Llama 2 Chat-HF} of 7 billion parameters (hereby referred to as \textit{Llama 7B}) \citep{touvron2023llama} to perform tuning experiments in this stage due to its manageable size and strong classification performance. These empirical findings then guide the final augmentation process.

\subsubsection{Optimization Procedure}\label{sec:prelim}
Below are the experiments we perform on a sample of the collected data to optimize these parameters.

\paragraph{Description of Procedure.} 
We choose 1,000 samples from each of the eight training datasets based on the following distribution: 20\% are selected from neutral samples (categories that do not indicate any offensive content, e.g., \textit{Not Hate, Normal}), while the remaining samples are evenly distributed among the non-neutral categories. Inspired by \citet{ho2022large} that diverse reasoning paths could improve knowledge distillation, we collect 4 alternative explanations, or \textit{degree of reasoning diversity}, generated by GPT-3.5-Turbo for these samples using temperature 0.7, resulting in 32,000 samples.

Figure \ref{fig:prompt_template} illustrates the \textit{Alpaca}-styled template to format each post with its corresponding label, generated explanation, definitions along with the instruction into blocks of an input prompt \citep{alpaca}. Using the described corpus, we supervised finetune \textit{Llama 7B} via LoRA techniques (technical details specified in Appendix \ref{apdx:tech}) \citep{hu2021lora}. Then, we perform zero-shot classification using the same template to prompt the finetuned model to generate the explanation and label for posts drawn from the test datasets \textit{HateCheck, HateXplain, Latent\_Hate}. We omit \textit{Implicit\_Hate} at this stage due to this set's markedly different 6-label space. For this part, posts are drawn using K-shot sampling (equal number of samples for each class) on the Validation portion of the test data, based on the values of \textit{K val} shown in Table \ref{tab:prelim_b}. 

\paragraph{Experiment Configurations.} For the first experiment, the training data is split into subsets whose distribution between the neutral (\textit{NE}) class and non-neutral class(es) (\textit{NN}) described by the following formula: $NN = R * NE $, where $R \in \{1, 2, 3, 4\} $ is the ratio coefficient. We set the number of explanations per sample to 2 \clt{, the smallest value that still enables the benefit of reasoning diversity}. For the second experiment, we construct the subsets by varying the \textit{degree of reasoning diversity} $D \in \{1, 2, 3, 4\} $ of each post. 
\paragraph{{Answer Extraction.}}\label{ans_extract} We extract the generated explanation and predicted labels after their respective tags. If the models generate multiple items from the dataset's label space, we select the first admissible label. If no acceptable output is obtained, we randomly select an item in the label space. 

\subsubsection{Insights and Augmentation}\label{sec:data_insight}
We report \textit{Llama 7B}'s macro F1-scores on the validation set of each configuration in Table \ref{tab:config}. A \textit{balanced distribution} between neutral and  non-neutral classes in the training corpus is beneficial, as reflected by the substantially high mean F1-score of 0.643 when R=1. On the other hand, having 3 explanations per sample (D=3) achieves the best overall performance across 3 test sets, consistent with \citet{ho2022large}'s findings on the benefit of multiple reasoning paths. However, performance markedly degrades when D=4. Our manual analysis reveals that the quality of generated outputs deteriorates as the degree of diversity increases, consequently affecting the performance of models trained on this data.

Guided by these empirical findings, we augment the training corpus by selecting approximately 1,800 extra samples from each of the 8 datasets while preserving the 1:1 balanced ratio of neutral to non-neutral classes. Then, we collect 3 explanations per sample using the described mechanism, resulting in a final corpus of 52,137 samples (Table \ref{tab:train_data}), hereby referred to as \textit{HateCOT}. 


\begin{table}[ht]
  \centering
  \begin{subtable}{\linewidth}
  \footnotesize
  \newcolumntype{R}[1]{>{\raggedleft\arraybackslash}p{#1}} 

    \centering
\begin{tabular}{lrrrr}
\hline
                                  & \textbf{R=1}            & \textbf{R=2}   & \textbf{R=3}   & \textbf{R=4}   \\ \hline
\multicolumn{1}{l|}{\citetalias{rottger-etal-2021-hatecheck} }    & 0.879          & 0.750  & 0.650  & 0.574 \\
\multicolumn{1}{l|}{\citetalias{mathew2021hatexplain}}   & 0.534          & 0.533 & 0.495 & 0.528 \\
\multicolumn{1}{l|}{\citetalias{elsherief2021latent}} & 0.516          & 0.473 & 0.456 & 0.408 \\ \hline
\multicolumn{1}{l|}{\textbf{Average} }    & \textbf{0.643} & 0.585 & 0.534 & 0.503 \\ \hline
\end{tabular}
    \caption{Results for Ratio configurations (R)}
    \label{tab:prelim_a}
  \end{subtable}
  \hfill
  \begin{subtable}{\linewidth}
  \footnotesize
    \centering
\begin{tabular}{lrrrr}
\hline
                                      & \textbf{D=1} & \textbf{D=2} & \textbf{D=3}   & \textbf{D=4} \\ \hline
\multicolumn{1}{l|}{\citetalias{rottger-etal-2021-hatecheck} }        & 0.851        & 0.879        & 0.864          & 0.783        \\
\multicolumn{1}{l|}{\citetalias{mathew2021hatexplain}}       & 0.549        & 0.534        & 0.597          & 0.607        \\
\multicolumn{1}{l|}{\citetalias{elsherief2021latent}}     & 0.480         & 0.516        & 0.477          & 0.465        \\ \hline
\multicolumn{1}{l|}{\textbf{Average}} & 0.627        & 0.643        & \textbf{0.646} & 0.618        \\ \hline
\end{tabular}
    \caption{Results for number of explanations per sample (D)}
    \label{tab:prelim_b}
  \end{subtable}
  \caption{Macro F1-scores for different configurations of distribution between neutral vs. non-neutral classes (top) and number of explanations per sample (bottom) on validation set. Best average performance in \textbf{bold}.}
  \label{tab:config}
\end{table}

%% file: experiment.tex
\section{Experiments on Test Sets}
We perform experiments to answer 3 questions. First, does \textit{HateCOT} improve zero-shot classification of open-sourced LLMs on unseen datasets? Second, how much data is necessary to enable competitive performance via in-domain finetuning after pretraining on \textit{HateCOT}? Finally, is in-context learning a viable alternative to finetuning?

\noindent{\textbf{Models}} In addition to \textbf{Llama 7B} in Section \ref{sec:prelim}, the following open-sourced models are selected.
\begin{itemize}[topsep=0pt, partopsep=0pt,itemsep=0pt, parsep=0pt,label=$\triangleright$]
    \item \textbf{Llama 13B} A larger variant of the instruction-tuned Llama 7B with 13 billion parameters. 
    \item \textbf{OPT-IML} Based on the original OPT (Open Pre-trained Transformer Language Models) \citep{zhang2022opt}, this encoder-only model contains 1.3 billion parameters and was further trained on the Instruction MetaLearning (IML) dataset \cite{iyer2022opt}.
    \item \textbf{Flan-T5-L} \citet{chung2022scaling} further instruction-finetuned the encoder-decoder T5 family of models \citep{raffel2020exploring}. We use the Large version of 780 million parameters. 
    \item \textbf{COT-T5-XL} A variant of the Flan-T5-XL (3 billion parameters), this model is further finetuned on the CoT dataset, a collection of 1.8 million samples augmented with chain-of-thought-style explanations \citep{kim2023cot}.
\end{itemize}

\subsection{Zero-shot Classification} We prompt the models to perform classification with no in-context examples via 2 modes: \textit{No Explanation}, where the model directly predicts the label for the input, and \textit{With Explanation}, where a justification is required before the predicted label. We finetune the base models using only \textit{HateCOT} and evaluate their performance on the 4 test sets as in Section \ref{sec:data_insight} (more details in Appendix \ref{apdx:tech}). 

\begin{figure*}[!ht]
  \centering

  \begin{subfigure}{\textwidth}
    \centering
    \includegraphics[width=\textwidth]{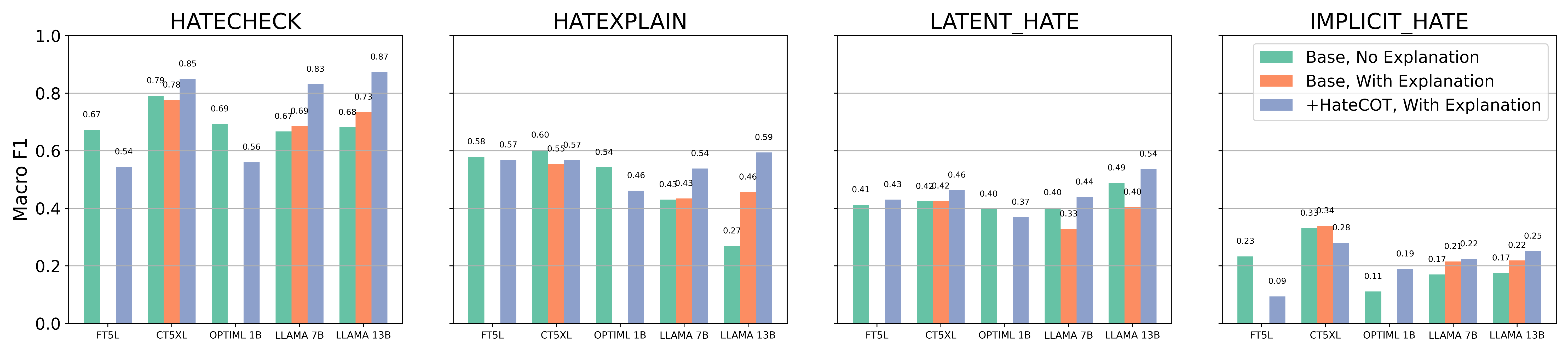}
    \caption{Macro F1 scores of LLMs in zero-shot setting using 3 configurations. \textit{Base} refers to out-of-the-box models, \textit{+HateCOT} denotes their pretrained counterpart on our dataset. \textit{FT5L}: Flan-T5-L, \textit{CT5XL}: COT-T5-XL. Results for Base Flan-T5-L and OPTIML models for \textit{With Explanation} settings omitted to reflect their inability to generate explanation according to the prompt.}
    \label{fig:zero_shot}
  \end{subfigure}%
  \hfill
  \begin{subfigure}{\textwidth}
    \centering
    \includegraphics[width=\textwidth]{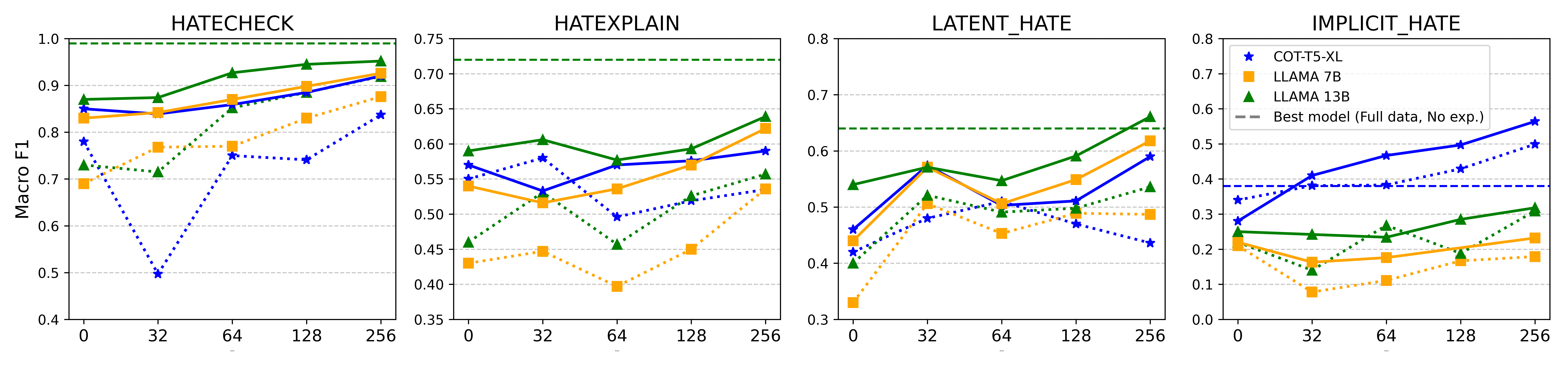}
    \caption{Macro F1 scores for models in zero-shot setting with explanation after K-shot in-domain finetuning at various values of K. Dashed line represents finetuned base models, solid line represents models pre-trained on \textit{HateCOT}, then in-domain finetuned. For each dataset, the horizontal dashed line represents the base version of the best performing model at K=256 which is finetuned using the entire training data \textit{without} any rationale for comparison, denoted as \textit{Best model (Full data, No. exp)}.}
    \label{fig:indomain}
  \end{subfigure}%
  \hfill
  \begin{subfigure}{\textwidth}
    \centering
    \includegraphics[width=\textwidth]{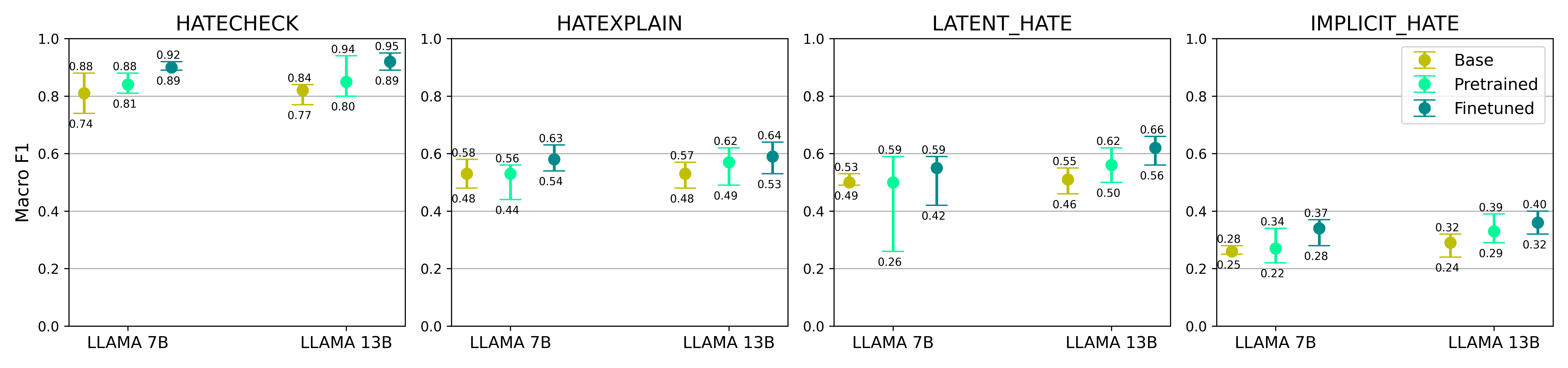}
    \caption{Min, max and mean of macro F1 scores for Llama 7B and Llama 13B over 5 seed using in-context learning. \textit{Pretrained} denotes models finetuned on \textit{HateCOT} only. \textit{Finetuned} denotes ICL performed on models that have been both \textit{HateCOT}-pretrained then K=256 shot in-domain finetuned.}
    \label{fig:icl}
  \end{subfigure}%

  \caption{Performance resutls of LLMs on test sets in various settings.}
  \label{fig:main}
\end{figure*}

From results presented in Figure \ref{fig:zero_shot}, the smaller models \textit{Flan-T5-L} and \textit{OPT-IML} are unable to generate explanations when prompted. In contrast, their scaled-up counterparts could follow instructions at all settings. Asking base (off-the-shelve) models to generate an explanation before the label results in observable boost to \textit{Llama} models on \textit{HateCheck} and \textit{HateXPlain}, but actually hampers performance on \textit{Latent\_Hate} and its derivative \textit{Implicit\_Hate}, which are notably challenging  due to its covert nature \cite{elsherief2021latent}. 

\paragraph{Model Choice Matters} Pretraining on \textit{HateCOT} unanimously enables all models to generate explanations. While smaller models receive no observable boost, larger models (\textit{COT-T5-XL, Llama 7B, Llama 13B}) are considerably enhanced compared to their base counterparts. With the exception of \textit{HateXplain}, the \textit{HateCOT}-pretrained version of \textit{COT-T5-XL} attains an increment in F1 scores of 7.6\% on \textit{HateXplain} and 9.5\% on \textit{Latent\_Hate} over the base counterpart without explanations. Similarly, \textit{Llama 7B} observes 23.9\%, 25.6\%, 10\% increment on \textit{HateCheck, HateXplain} and \textit{Latent\_Hate}, respectively. These statistics are 27.9\%, 118.5\%, and 10.2\% for \textit{Llama 13B}. Notably, all models yield non-competitive performance  on \textit{Implicit\_Hate}.

The reduced performance of \textit{HateCOT}-pretrained models compared to their base counterparts without explanations (e.g.: \textit{Flan-T5-L} and \textit{OPT-IML} on almost all test sets) is in line with literature as COT-style prompting tends to  favor larger models \cite{wei2022chain, wang2024exploring, suzgun2023challenging}. Even the reduced F1 scores of \textit{COT-T5-XL} on \textit{HateXPlain} and \textit{Implicit\_Hate} is consistent with this model's suboptimal performance relative to its larger variants, as showcased in \citet{kim2023cot}. These results serve as an empirical reference for researchers to select the appropriate model size for their respective task. 

\subsection{In-Domain Finetuning}
We further finetune \textit{COT-T5-XL, Llama 7B, Llama 13B} using data from the training portions of the test datasets, including \textit{Implicit\_Hate}. To simulate low-resource settings, we choose 256 samples uniformly at random from each class, then augment them with explanations as described in Section \ref{sec:exp}.  Both the \textit{Base} and \textit{Pretrained} versions of the models are then finetuned using various K-shot $\in \{32,64,128,256\}$ training data from this pool.

In Figure \ref{fig:indomain}, the general superiority of finetuning models after \textit{HateCOT} over their base counterpart indicates enhanced generalizability with limited in-domain data. However, too little training data (K$\leq$64) may impair models' performance compared to the zero-shot setting, likely a result of attempting to optimize a large number of parameters on limited signals. Stable gains are attained at K=128, and at K=256, significant boost over the non-finetuned zero-shot results are observed.

Interestingly, decoder-only models (\textit{Llama}) considerably outperform encoder-decoder \textit{COT-T5-XL} on the 2 and 3-way classification tasks, yet the reverse is observed for the nuanced 6-way \textit{Implicit\_Hate}. On this task, only \textit{COT-T5-XL} consistently scales with the increment in training data to reach the max F1 score of 0.56, while \textit{Llama} models plateau at sub-0.3 range even at K=256. 

We further select the best performing model at K=256 for each dataset and finetune their \textit{Base} versions with the entire training data and no explanations no definition. In Figure \ref{fig:indomain} and Table \ref{apdx:indomain}, in-domain finetuning after \textit{HateCOT} achieves competitive results even with only a fraction of the full training data. Furthermore, prompting for explanations enables \textit{Llama 13B} and \textit{COT-T5-XL} to attain performance that surpasses using the full training data on \textit{Latent\_Hate} and \textit{Implicit\_Hate}.

\subsection{In-context Learning}
As an alternative to in-domain finetuning, we investigate the models' performance using in-context learning (ICL), when a number of complete examples are provided as part of the input prompt. We select 1 sample from \textit{each} class in the training data of each dataset, then obtain its the associated explanation from GPT-3.5-Turbo. The sets of post, explanation and label are arranged in the same format in the same template shown in Figure \ref{fig:prompt_template}. We run inference for classification results over 5 seeds, which also randomly permutes their order. 

Figure \ref{fig:icl} shows the mean, minimum and maximum values of macro F1 scores over the seeds for the base, \textit{HateCOT}-pretrained only (\textit{Pretrained)}, and in-domain finetuned (K=256) versions of \textit{Llama 7B} and \textit{Llama 13B}. \textit{COT-T5-XL} regularly generates overly repetitive outputs, and thus omitted. The range of F1 scores is large regardless of settings, an observation in line with  the variance of in-context learning in literature \cite{lu2022fantastically, dong2022survey}. Unsurprisingly, base models' performances tend to be inferior to their finetuned counterpart. Interestingly, the max F1 scores of finetuned models with ICL are not appreciably better than those in the zero-shot counterparts (Figure \ref{fig:indomain}). In contrast, except for \textit{Llama 7B} on \textit{HateXplain}, the best scores of pretrained models approach those of the finetuned models--particularly for \textit{Llama 13B}.

This finding suggests another advantage of pre-training on \textit{HateCOT}: boosting performance via ICL without additional in-domain finetuning, an area that has attracted growing attention \cite{min2021metaicl, wang2023large, ye2023context}.  Nevertheless, there exists the trade-off: ICL examples with explanations extend significantly the context length, and ICL inferencing takes considerably more time compared to zero-shot, making the latter more resource-efficient overall. 

\begin{table*}[!htb]
\centering
\footnotesize
\begin{tabularx}{0.8\textwidth}{@{}Xlccrrrr@{}}
\toprule
\makecell[ll]{\textbf{Dataset}} & \makecell[cc]{\textbf{Best Model} \\ \textbf{@K=256}} & \makecell[cc]{\textbf{F1} \\ \textbf{@K=256}} & \makecell[cc]{\textbf{F1 Base} \\ \textbf{+ Full}} & \makecell[rr]{\textbf{F1 \%}} & \makecell[rr]{\textbf{Data Size} \\ \textbf{@K=256} } & \makecell[rr]{\textbf{Data Size} \\ \textbf{Full}} & \makecell[rr]{\textbf{Data \%}} \\
\midrule
\textit{HateCheck}        & LLAMA 13B                     & 0.95                  & 0.99                     & 96\%              & 512                     & 1,864                      & 27\%                  \\
\textit{HateXplain}       & LLAMA 13B                     & 0.64                  & 0.72                     & 89\%              & 768                     & 12,088                     & 6\%                 \\
\textit{Latent\_Hate}     & LLAMA 13B                     & 0.66                  & 0.64                     & 103\%             & 768                     & 11,460                     & 7\%                   \\
\textit{Implicit\_Hate}   & COT-T5-XL                       & 0.56                  & 0.38                     & 147\%             & 1,536                    & 2,707                      & 57\%                  \\
\bottomrule
\end{tabularx}
\caption{Comparison of performance metrics for the best performing models finetuned using K=256 in-domain post \textit{HateCOT} vs. finetuned on the full training set and no explanation nor definition. \textit{F1\%} denotes the percentage of macro F1 score of the K=256 finetuned model over that of the model trained on full data. Similarly, \textit{Data \%} denotes the percentage of data size used by the K=256 regimen over the full data.}
\label{apdx:indomain}
\end{table*}

\subsection{Assessment and Recommendations}

From empirical observations, we make the following recommendations to construct a cost-efficient pipeline for classifier on novel domains: 
\begin{itemize}[topsep=0pt, partopsep=0pt,itemsep=0pt, parsep=0pt,label=$\triangleright$]

\item The most consistent benefit of \textit{HateCOT} is its capacity to enable data-efficient in-domain finetuning following pretraining.

\item Practitioners should choose models of sufficient number of parameters for the task. Larger instruction-tuned LLMs appear to more effectively capitalize on \textit{HateCOT} pretraining regimen before in-domain finetuning.

\item Instead of devoting resources to curate substantial training data, practitioners could focus on obtaining high quality annotations for representative rationales, and augment them into explanations using their LLM of choice. Alternatively,  practitioners may choose to curate the explanations organically to achieve certain desired thematic qualities. This process may be iterated until targeted performance is reached according to some guiding metrics with acceptable quality of explanation.
\end{itemize}


%% file: qa.tex
\section{Quality of Explanations}
In addition to the enhanced classification prowess, we investigate whether pretraining on \textit{HateCOT} also improves the quality of explanation LLMs. To this end, the following 2 human quality assurance (QA) experiments are conducted. In QA 1, we assess if human annotators prefer the explanations generated by the \textit{base} or \textit{HateCOT}-pretrained LLMs (\textit{COT-T5-XL, Llama 7B, Llama 13B}). In QA 2, we perform in-domain K-shot finetuning on the aforementioned models and examine how annotators evaluate their generated explanations.  An additional assessment of Target Identification is presented in Appendix \ref{qa3}.

\subsection{QA 1: Base vs. Pretrained}\label{sec:qa1}
From the outputs of the test sets generated by the 3 LLMs, we select 50 samples uniformly at random whose explanations from the \textit{Base} and \textit{HateCOT}-pretrained versions agree on the predicted label, for a total of 150 samples and 300 explanations. We then recruit 13 annotators from the crowdsource platform Amazon Mechanical Turk and solicit their annotation on these explanations (Appendix \ref{qa_val}).  Using the template in Figure \ref{apdx:qa_1}, we ask the annotators to indicate their preferred explanation that better suits the purpose of content moderation based on fluency, soundness and the alignment with the definition of the chosen label. Each post is annotated by 5 humans, resulting in 750 annotations.

In Table \ref{tab:qa_1}, we observe that the raw frequency count for explanations generated by the \textit{HateCOT}-pretrained models exceed their base version's. Similarly, even when tallying by majority vote--where the explanation is chosen by at least 3 out of 5 annotators--preference for those generated by the \textit{Pretrained} models still prevails. We note that the preference margin is smaller for \textit{Llama 13B} \textit{Pretrained}, likely due to this model's already strong generative capabilities.

\begin{table}[t]
\resizebox{\columnwidth}{!}{%
\begin{tabular}{lrrrr}
\toprule
 & \multicolumn{2}{c}{\textbf{Human (Frequency Count)}} & \multicolumn{2}{c}{\textbf{Human (Majority Vote)}} \\
\cmidrule{2-5}
\textbf{Model} & \textbf{Base} & \textbf{Pretrained} & \textbf{Base} & \textbf{Pretrained} \\
\midrule
COT-T5-XL & 62 (24.8\%) & 188 (75.2\%) & 6  (12\%) & 44 (88\%) \\
Llama 7B & 109 (43.6\%) & 141 (56.4\%) & 19 (38\%) & 31 (62\%) \\
Llama 13B & 114 (45.6\%) & 136 (54.4\%) & 22 (44\%) & 28 (56\%) \\
\bottomrule
\end{tabular}%
}
\caption{Comparison of Base and Pretrained models in Human Evaluation. \textit{Frequency Count} : count per annotation; \textit{Majority Vote} indicates aggregate count by the version is preferred by at least 3 out of 5 annotators.}
\label{tab:qa_1}
\end{table}

\subsection{QA 2: Inter-model Comparison}
Inspired by \citet{wang2023evaluating, lin2023unlocking, yang2023hare}, we assess the quality of explanations generated by finetuned models (K=256) on the following criteria: 
\begin{itemize}[topsep=0pt, partopsep=0pt,itemsep=0pt, parsep=0pt,,label=$\triangleright$]
    \item \textbf{Persuasiveness}: how convincingly the explanation justifies its chosen label for the post. 
    \item \textbf{Soundness}: how valid and logical is the explanation with respect to the label's definition.
\end{itemize}

\begin{figure}[!hbt]
  \centering
    \includegraphics[width=\columnwidth]{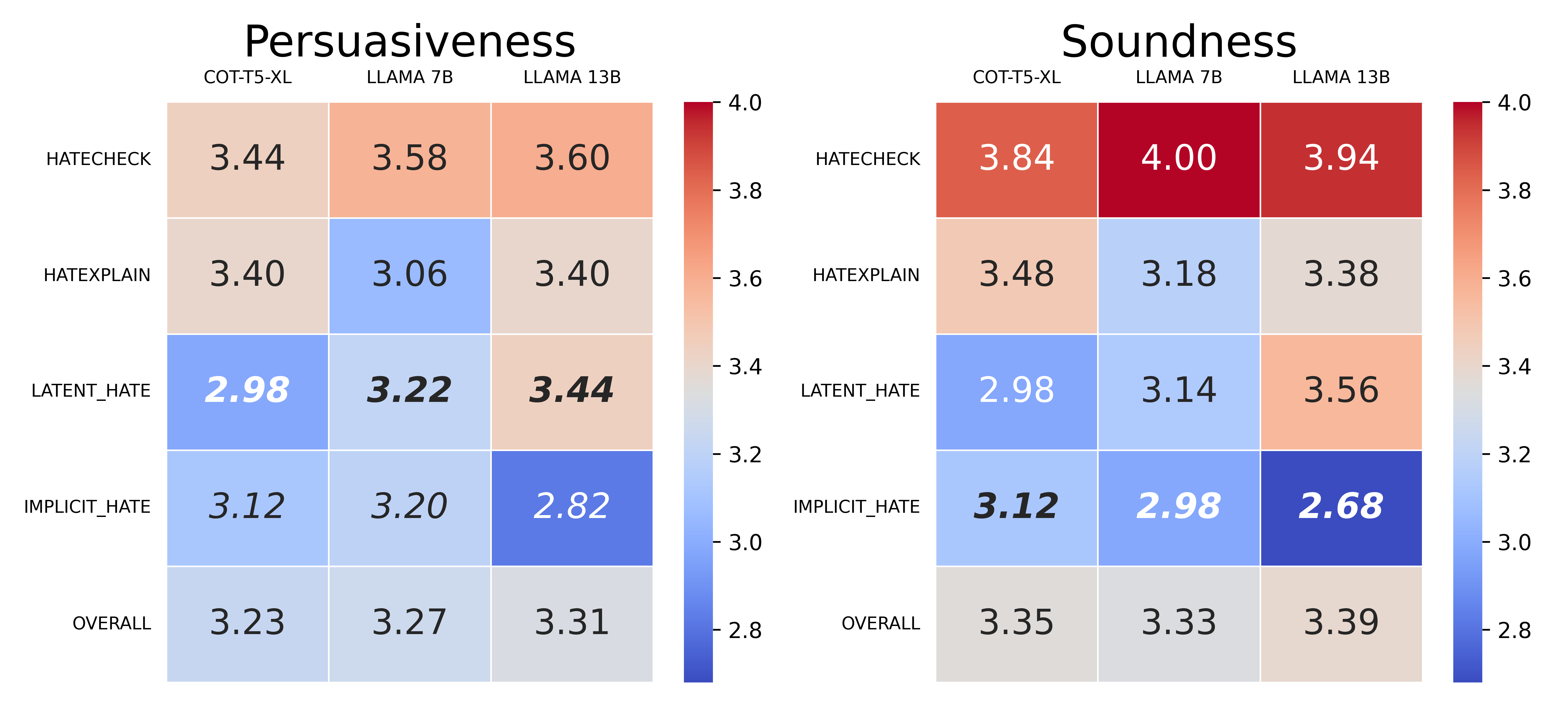}
    \caption{Heatmap for the average rating of explanations by finetuned \textit{Model} (x-axis) and \textit{Dataset} (y-axis) on 3 criteria from 1 (least) to 5 (very). \textit{Overall} indicates average scores aggregated over all datasets. Triplets of scores \textit{italicized} and in \textbf{bold} are those whose $p$-value < 0.05 by one-way ANOVA test that compare ratings of 3 models across the dataset on that row. \textit{Italicized}-only scores indicate marginal significance ($p$-value $\approx$ 0.07).}
    \label{fig: qa_2}
  \end{figure}

For this QA task, we recruit 6 annotators also from Amazon Mechanical Turk (Appendix \ref{qa_val}). Using the template in Figure \ref{apdx:qa_2}, we collect their numerical ratings on a scale from 1 (least) to 5 (very) on these criteria for 50 posts per model per dataset, for a total of 600 annotations. Figure \ref{fig: qa_2} displays the mean ratings for each model-dataset pair, as well as \textit{Overall} scores averaged across all datasets.
 The average \textit{Ovearall} ratings for both \textit{Persuasiveness} and \textit{Soudness} are above 3.2 out of 5, indicating generally positive reception by human evaluators. Interestingly, there exists a degree of correlation between the models' better classification performance (Figure \ref{fig:indomain}) and higher mean ratings on each dataset, a useful artifact to calibrate models. These ratings may serve as benchmarks for future works. 

\section{Conclusion} 
We show empirically that our \textit{HateCOT} dataset considerably enhances offensive speech detection even with limited training data while producing high-quality explanations. We invite future research to explore other benefits of LLM-augmented data and extend them to other related low-resource areas.

%% file: limitation.tex
\section*{Limitations}
\clt{We first acknowledge that our work is restricted to English corpora, a common limitation among literature on offensive speech \cite{yin2021towards, poletto2021resources}. However, our approach sets a proof-of-concept for researchers to construct similar corpus in other languages by leveraging existing resources. Furthermore, our developmental pipeline is considerably more data-efficient than conventional approaches (Section 4.1, 4.2), potentially lowering the barrier of entry for practitioners without access to abundant resources. Therefore, this work invites further expansion on multilingual datasets, particularly to develop corpora with clearly defined definitions to facilitate synergy with other research. }

Second, due to computational limitations, we could not perform experiments on larger open-source models. With the development of newer, more powerful models, it is reasonable to expect their performance to further improve though the use of this dataset of our corpus, as demonstrated by our empirical results. 

Finally, we recognize the risk of propagating implicit biases that LLMs are known to carry \cite{cheng2023marked, gupta2023bias}. However, we note that the approach of using LLMs (GPT-3.5-Turbo in this paper) to bridge the logical gaps in original rationales has been shown to produce outputs less prone to logical failures \cite{yang2023hare}. Biases in Pretrained Language Models have been attracting much attention in the research community. We invite further works to consider our approach to reduce hallucinations and biases in text generation.

\section*{Ethics Statement}
 We acknowledge the potential malicious usage of our corpus to generate content capable of evading detection, or jeopardizing classifiers' performance. 

 \section*{Acknowledgement}
 This work is funded by the University of Maryland's Institute for Trustworthy AI in Law \& Society (TRAILS). We thank the service of ACL ARR reviewers, area chairs and the editors of the EMNLP conference for our paper's publication.

%% file: appendix.tex
\onecolumn
\appendix
\section{Appendix}
\label{apdx}
\subsection{Data Pre-Processing}
The datasets used in this work are released by their respective authors for research purpose. Aware of the risk of containing confidential in social media data, we anonymize posts during the curation process by replacing user handles with the string '<user>'. Our many layers of randomization provides further protection with respect to privacy.

\subsection{Technical Specification}\label{apdx:tech}
\subsubsection{Inferencing}
We used OpenAI's API to access the publicly available version of GPT-3.5 in November 2023, and GPT-4 in January 2024. To obtain explanations from the former (as described in \ref{sec:data_insight}), we choose temperature among candidates $\{0.3, 0.5, \textbf{0.7}\}$ and settle on the last value during inference. This value is selected based on literature and multiple iterations of qualitative analysis of outputs  \cite{yang2023hare, kim2023cot}. For GPT-4 and other open-source models, we use greedy decoding. 

\subsubsection{Finetuning}
To train models, we employ both full supervise finetuning (\textit{FLAN-T5-Large, OPT-IML}) and LoRA parameter-efficient techniques (all other models). LoRA models set to 8-bit quantization using the BitsandByes library. Training \textit{FLAN-T5-Large, OPT-IML, LLAMA 7B} models was done on 2 Nvidia RTX A6000 GPUs, whereas \textit{COT-T5-XL, Llama 13B} used 4 GPUs. Hyperparameters for the following candidates are tuned on the validation set of sampling K=64 shots from the leftover training samples by optimizing macro F1-score metric. Options in bold indicate final chosen values among multiple across all models to finetune on \textit{HateCOT}.
\begin{itemize}[topsep=0pt, partopsep=0pt,itemsep=0pt, parsep=0pt]
    \item Learning rate: \{5e-5, 1e-4, \textbf{3e-4}\}
    \item Training Epochs: \{1, 2, \textbf{3}\}
    \item LoRA Rank: \{16, 32, \textbf{64}\} (\textit{alpha}$=$rank*2)
    \item LoRA Target Modules: \{Q, V\}
    \item Batch size: 2 
    \item Gradient Accumulation Step: 2
\end{itemize}
For in-domain K-shot finetuning,  the values above remain the same except for the following variations in Learning Rates, which is set to 1e-4 for \textit{HateCheck, HateXplain, Latent\_Hate}, and 3e-4 otherwise.

\subsection{Variants of Prompt Template for Explanation}\label{apdx:instruction}
\begin{figure*}[t]
  \centering
  \includegraphics[width=0.6\textwidth]{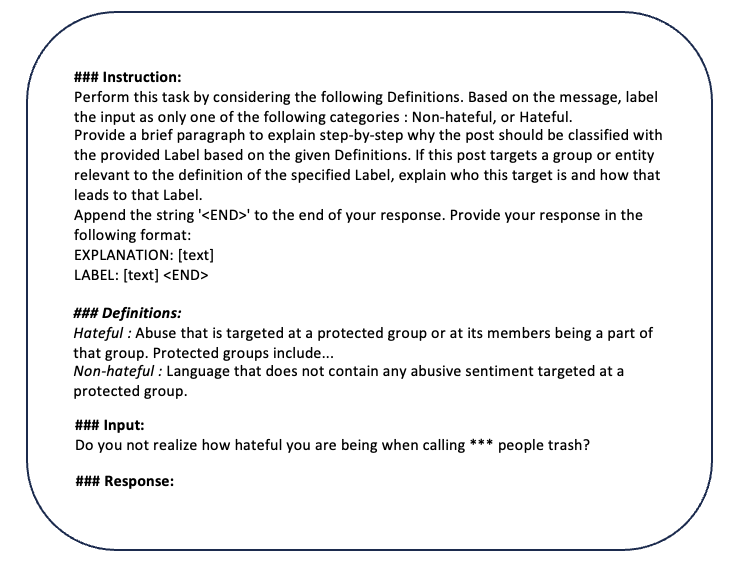}
  \caption{Template used to prompt LLM for classification inference.}
  \label{fig:prompt_template}
\end{figure*}

The 8 datasets introduced in Section \ref{sec:data} may not always have annotations on all required fields; thus, we modify the first sentence in the \textit{Instruction} block in Figure \ref{fig:prompt_template} with the following variants when appropriate: 
\begin{itemize}[topsep=0pt, partopsep=0pt,itemsep=0pt, parsep=0pt]
    \item Only \textit{Target} is available: \textit{'Provide a brief paragraph to explain step-by-step how the post targets the specified group or entity, and how that leads to the specified Label based on the given Definitions.'}
    \item Both \textit{Target} and \textit{Rationale} are available: \textit{'By elaborating on the provided Annotation, provide a brief paragraph to explain step-by-step how the post targets the specified group or entity, and how that leads to the specified Label based on the given Definitions.'}
\end{itemize}

\subsection{Quality Review of Sample Explanations }\label{apx:manual_review} 
Elaborating on the criteria outlined in Section \ref{sec:exp}, we review the quality of the generated explanation by GPT-3.5-Turbo on the following items: 
\begin{itemize}[topsep=0pt, partopsep=0pt,itemsep=0pt, parsep=0pt]
    \item Grammatically correct
    \item Succinct in their justification of the chosen label.
    \item Persuasive and logical in their reasoning for the chosen label
\end{itemize}
We discard explanations that are too verbose, and/or not choosing the label already provided by human annotation (fortunately, this scenario happens rarely, likely due to the presence of existing rationales guiding the extra generated outputs). We also remove explanations that conjure incorrect and/or irrelevant facts to the context to discourage hallucination and encourage a high degree of pertinent to the target post.

\begin{table}[!t]
\centering
  \begin{tabularx}{0.55\textwidth}{@{}lccc@{}}
\toprule
\textbf{Dataset} & \textbf{COT-T5-XL} & \textbf{Llama 7B} & \textbf{Llama 13B} \\
\midrule
\textit{HateCheck} & 94.5 \% & 96.5 \% & 93.5 \% \\
\textit{HateXplain} & 71.0 \% & 74.0 \% & 74.0 \% \\
\textit{Latent\_Hate} & 50.5 \% & 51.0 \% & 44.5 \% \\
\bottomrule
\end{tabularx}
\caption{Percentage out of 200 samples per dataset,  where explanations correctly identify at least 1 of the targets listed by human annotators by each model.}
\label{tab:target}
\end{table}

\subsection{Human Annotation for QA Experiment}\label{qa_val}
\subsubsection{QA 1}
With approved IRB, we recruit 13 crowdsource workers using the Amazon Mechanical Turk platform to annotate 50 samples per model, for a total of 150 data points for the task described in Section \ref{sec:qa1}. The annotators was paid a fair wage at \$15 per hour, and forewarned about the nature of the task. Annotators must be fluent English speakers. We also limit each annotator to no more than 100 posts (60\% of the total 150 samples per model) to maintain diversity of opinions. We observe that preference for explanations generated by \textit{Pretrained} models remains consistent with GPT-4's. 

The demographic breakdown of the 13 annotators are described below: 
\begin{itemize}
    \item \textbf{Gender}: Female (8), Male (5)
    \item \textbf{Age}: 18-29 (2), 30-39 (4), 40-49 (4), 50+ (3)
    \item \textbf{Education}: High School (2), 2-year college (5), 4-year college (4), Master's or Higher (2)
\end{itemize}

\subsubsection{QA 2}
For this experiment, each annotation task consists of the explanations generated by 3 models are grouped by the sample. This division resutls in 200 tasks. 6 Amazon Mechanical Turk workers are recruited, with similar qualification criteria as described above.

The demographic breakdown of the 6 annotators are described below: 
\begin{itemize}
    \item \textbf{Gender}: Female (3), Male (3)
    \item \textbf{Age}: 18-29 (1), 30-39 (2), 40-49 (3)
    \item \textbf{Education}:  2-year college (2), 4-year college (4)
\end{itemize}

\subsection{QA 3: Target Identification}\label{qa3}
\begin{figure*}[!t]
  \centering
    \includegraphics[width=0.6\textwidth]{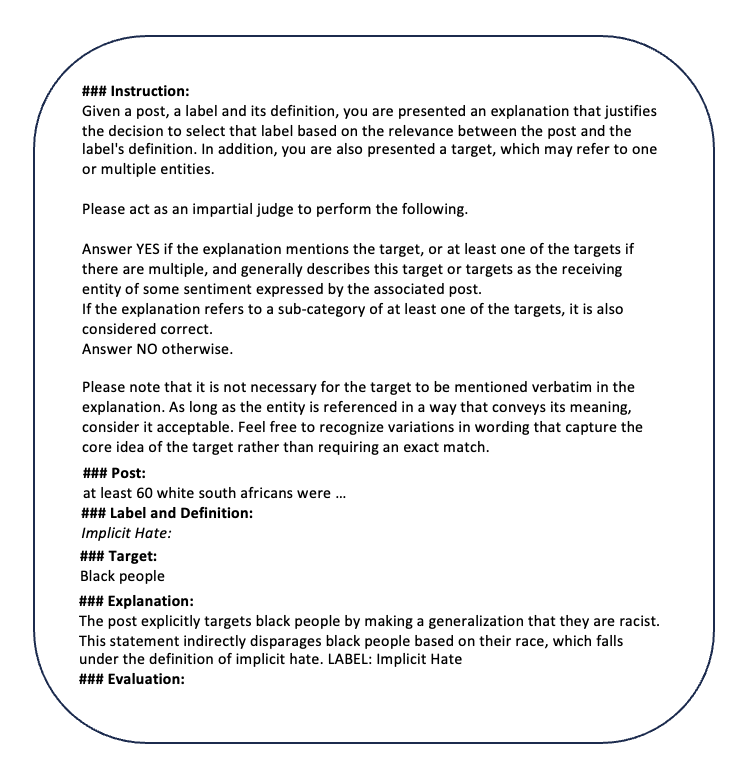}
  \caption{Template for QA Experiment 3.}
  \label{apdx:qa_3}
\end{figure*}

We investigate the in-domain finetuned models' capabilities to identify the target of the sentiments expressed by the post. We also randomly select 200 samples from each dataset that have the \textit{Target} variable annotated by humans, then ask GPT-4 to judge whether the explanations from the models mention at least one of the listed targeted groups using the template in Figure \ref{apdx:qa_3}. Note that we encourage GPT-4 to consider potential variance of expression and not restrict to exact matches.

Table \ref{tab:target} shows the percentage of accuracy on this task. Similar levels of descending performances are observed in the presented order of test datasets. However, this observation may be an artifact of the differences between the annotations targets among datasets: \textit{HateCheck} has a limited number of discrete categories while \textit{Latent\_hate} contains multiple combinations of free-text labels. We urge practitioners to consider this factor while curating training data if target identification is desired.

\begin{table*}[hp]
\small
\begin{center}
\begin{tabular}{p{2.5cm}p{12cm}}
\toprule  \textbf{Dataset} &  \textbf{Definition} \\
\midrule 
\hline
\citealp{salminen2018anatomy} & \textbf{Neutral} : A post that is not offensive to any group of people. \textbf{Hateful}: An offensive post, motivated, in whole or part, by the writer’s bias against an aspect of a group of people.\\ \hline

\citealp{qian2019benchmark}  & \textbf{Not Hate} : Does not contain any attack on people based on protected characteristics.  \textbf{Hate}: Contains direct attack on people based on protected characteristics such as race, ethnicity, national origin, religious affiliation, sexual orientation, caste, sex, gender, gender identity, and serious disease or disability. \\ \hline

\citealp{sap2019social} & \textbf{ Not Offensive} : not offensive to anyone.  \textbf{Offensive} : denotes the overall rudeness, disrespect, or toxicity of a post. whether a post could be considered offensive to anyone. \\ \hline

\citealp{vidgen2021introducing} & \textbf{Neutral}: Content that does not fall into other categories, usually entirely unrelated to abuse, hate, prejudice, or intolerance. \textbf{Identity Directed Abuse}:  Content that directs abuse at an identity, which relates to fundamental aspects of individuals’ social position, community and self-representation. An identity includes but is not limited to religion, race, ethnicity, gender, sexuality and sexual preferences, immigration status, nationality, ableness, physical appearance and class.  \textbf{Affiliation Directed Abuse} : Content that directs abuse at people who have a voluntary affiliation with a profession, membership, association, ideology, or other well-defined group or collective. \textbf{Person Directed Abuse} : Content that directs abuse at an identifiable person. \\ \hline

\citealp{vidgen-etal-2021-learning}  &  \textbf{Derogation} : content which explicitly attacks, demonizes, demeans or insults a group. \textbf{Animosity} : content which expresses abuse against a group in an implicit or subtle manner. \textbf{Threatening} : content which expresses intention to, support for, or encourages inflicting harm on a group, or identified members of the group. \textbf{Support For Hateful Entities} : content which explicitly glorifies, justifies or supports hateful actions, events, organizations, tropes and individuals collectively, entities. \textbf{Dehumanization} : content which perceives or treats people as less than human. \textbf{Not Hate} : content that falls into none of the other categories. \\ \hline

\citealp{basile2019semeval} &  \textbf{Hate Speech} : language that is used to expresses hatred towards a targeted group or is intended to be derogatory, to humiliate,or to insult the members of the group. may also be language that threatens or incites violence. \textbf{Offensive Language} : may contain offensive terms but targets disadvantaged social groups in a manner that is potentially harmful to them. \textbf{Neither} : language that does not all into either of the other categories.\\ \hline

\citealp{hartvigsen2022toxigen} & \textbf{Benign}: Text that is not harmful nor offensive to anyone. \textbf{Toxic}: Text that could be seen as harmful to anyone, may contain offensive, rude humor, insults, personal attacks, profanity, aggression, may refer to targeted group with harmful intent that is expressed in stereotypes or lewd manners\\ \hline

\citealp{toraman2022large}  &  \textbf{Hate} : target, incite violence against, threaten, or call for physical damage for an individual or a group of people because of some identifying trait or characteristic. \textbf{ Offensive} : humiliate, taunt, discriminate, or insult an individual or a group of people in any form, including textual. \textbf{Normal} : does not fall into any of the other categories.\\
\bottomrule
\end{tabular}
\caption{Labels and definitions of 8 datasets used to synthesize training corpus. Definitions are lifted directly from their original works, with exceptions of minor adjustment for stylistic consistency.}
\label{apx:train_def}
\end{center}
\end{table*}

\begin{table*}[!ht]
\footnotesize
\begin{center}
\begin{tabular}{p{2.5cm}p{12cm}}
\toprule  \textbf{Dataset} &  \textbf{Definition} \\ \hline  \hline
HateCheck & \textbf{Non-hateful}: Language that does not contain any abusive sentiment targeted at a protected group. \textbf{Hateful}: Abuse that is targeted at a protected group or at its members being a part of that group. Protected groups include age, disability, gender identity, familial status, pregnancy, race, national or ethnic origins, religion, sex or sexual orientation. \\  \hline
HateXplain & \textbf{Normal} : neither hate speech nor offensive. \textbf{Hate Speech} : language which attacks, demeans, offends, threatens, or insults a group based on race, ethnic origin, religion, disability, gender, age, sexual orientation, or other traits. it is not the presence of certain words that makes the text hate speech, rather you should look the context the word is used in the text. \textbf{Offensive Language} : usage of rude, hurtful, derogatory, obscene or insulting language to upset or embarrass people. \\ \hline
Latent\_Hate &  \textbf{Not Hate} : speech or actions that do not involve any form of hatred, prejudice, or discrimination toward individuals or groups based on their characteristics. \textbf{Explicit Hate}: openly expressed, direct forms of hatred and prejudice toward individuals or groups based on their characteristics. \textbf{Implicit Hate}: coded or indirect language that disparages a person or group on the basis of protected characteristics like race, gender, and cultural identity.\\ \hline
Implicit\_hate & \textbf{White Grievance} : includes frustration over a minority groups perceived privilege and casting majority groups as the real victims of racism. This language is linked to extremist behavior and support for violence. \textbf{Incitement To Violence} : includes flaunting in group unity and power or elevating known hate groups and ideologies. \textbf{Inferiority Language }: implies one group or individual is inferior to another, and it can include dehumanization, denial of a person's humanity , and toxic language that compares the target with disease, insects, animals . Related to assaults on human dignity, dominance, and declarations of superiority of the in group. \textbf{Irony} : refers to the use of sarcasm , humor, and satire to attack or demean a protected class or individual. \textbf{Stereotypes And Misinformation} : associate a protected class with negative attributes such as crime, or terrorism. includes misinformation that feeds stereotypes and vice versa, like holocaust denial and other forms of historical negationism. \textbf{Threatening And Intimidation }: conveys a speaker's commitment to a target's pain, injury, damage, loss or violation of rights, threats related to implicit violation of rights and freedoms, removal of opportunities, and more subtle forms of intimidation. \\ 
\bottomrule
\end{tabular}
\end{center}
\caption{Labels and definitions for 4 test datasets.}
\label{apx:test_def}
\end{table*}

\begin{figure*}[t]
    \centering
    \begin{subfigure}{\textwidth}
        \centering
        \includegraphics[width=\textwidth]{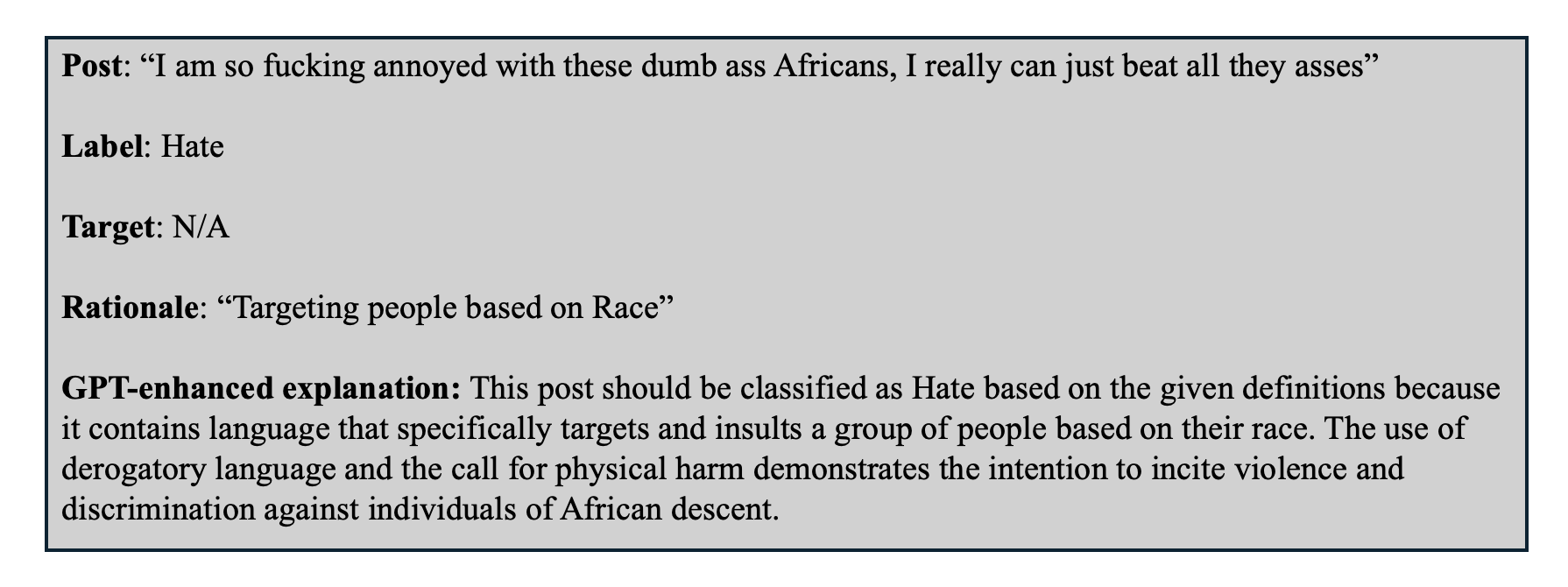}
        \caption{Example from \citet{toraman2022large}}
    \end{subfigure}
    \begin{subfigure}{\textwidth}
        \centering
        \includegraphics[width=\textwidth]{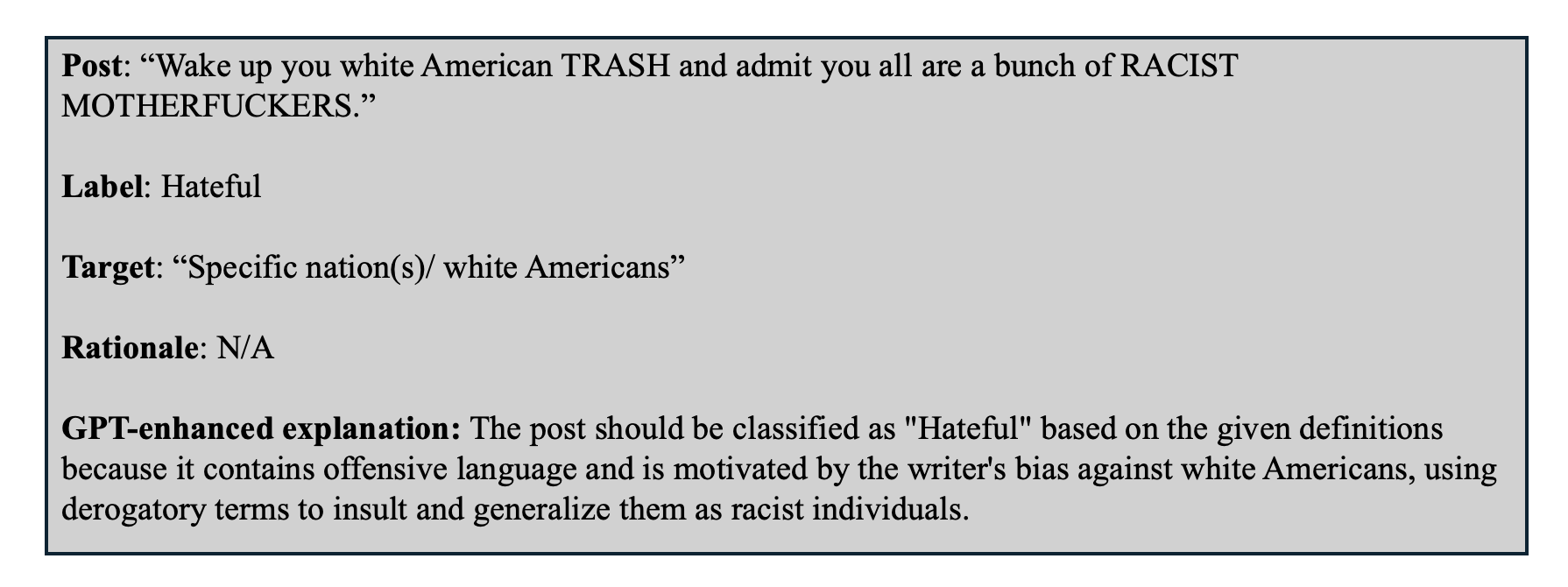}
        \caption{Example from \citet{salminen2018anatomy}}
    \end{subfigure}
    \begin{subfigure}{\textwidth}
        \centering
        \includegraphics[width=\textwidth]{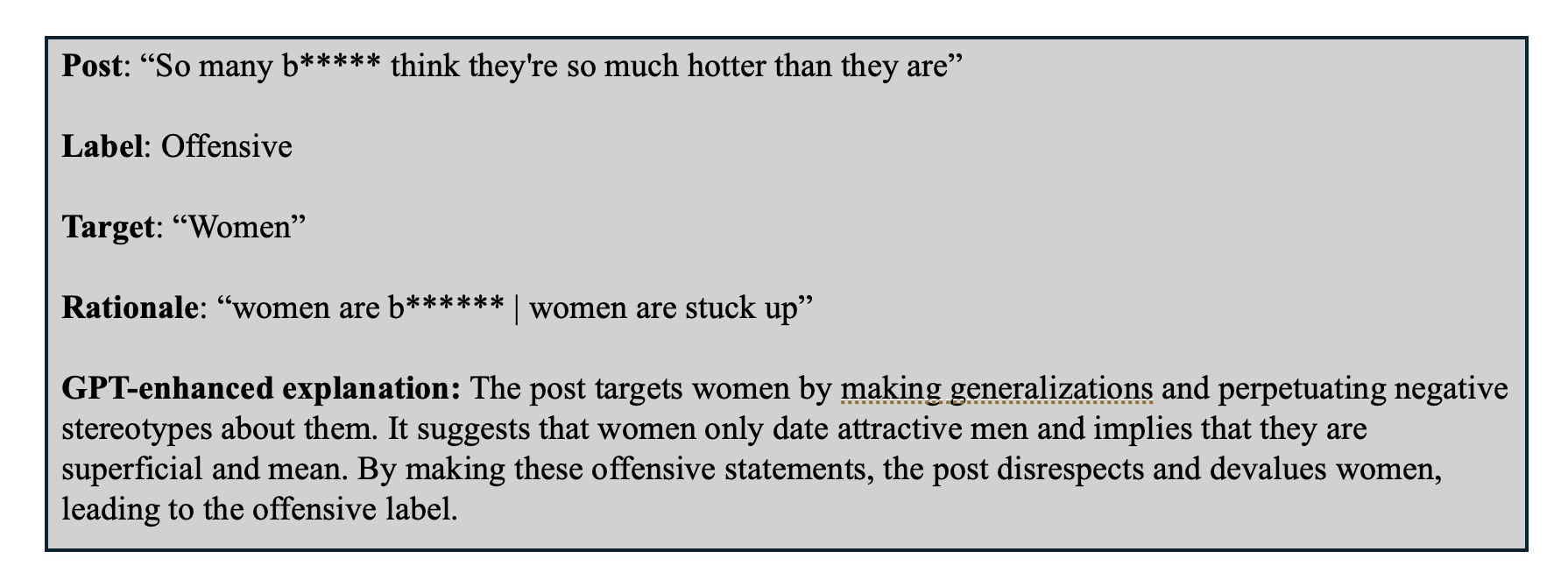}
        \caption{Example from \citet{sap2019social}}
    \end{subfigure}
    \caption{Examples drawn from our training corpora showing their native \textit{Post, Target and Rationale}, along with the corresponding GPT-3.5-Turbo-enhanced explanations. Due to their nature as fragmented annotations, verbatim \textit{Rationales} are not serviceable explanation, but can serve as guiding signals that leverage GPT's generative capabilities to construct legible passages with detailed justifications.}
    \label{fig: rationale}
\end{figure*}

\begin{figure*}[!t]
  \centering
    \includegraphics[width=0.6\textwidth]{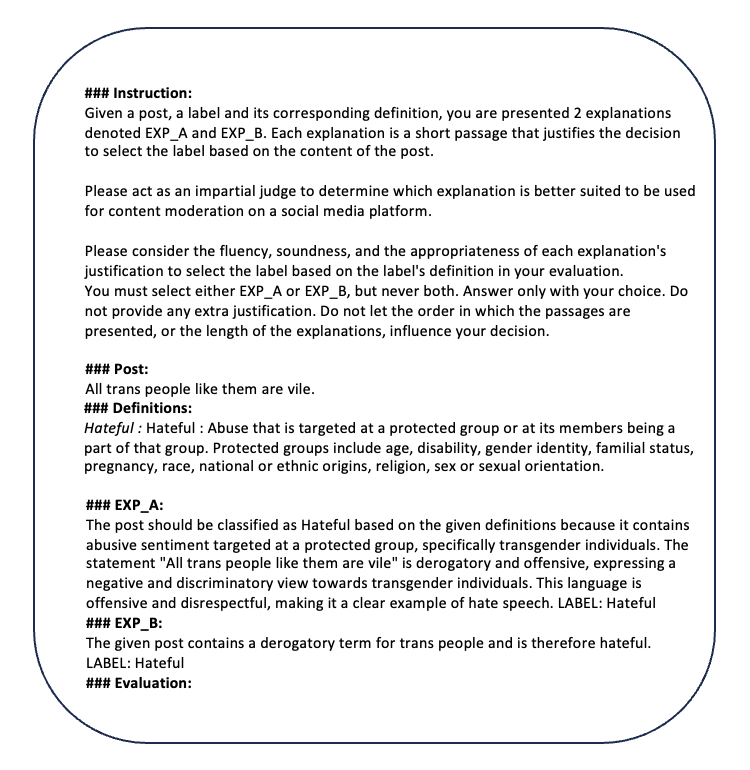}
  \caption{Template for QA Experiment 1. In this example, EXP\_A is preferred.}
  \label{apdx:qa_1}
\end{figure*}

\begin{figure*}[!t]
  \centering
    \includegraphics[width=0.6\textwidth]{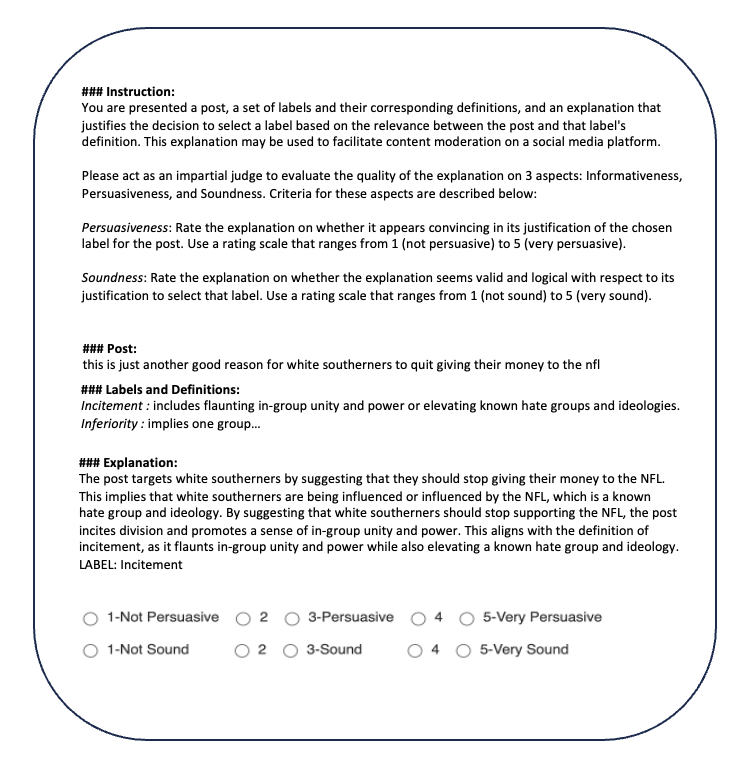}
  \caption{Template for QA Experiment 2.}
  \label{apdx:qa_2}
\end{figure*}

